\documentclass[10pt,twocolumn,letterpaper]{article}

\usepackage{cvpr}              
\definecolor{cvprblue}{rgb}{0.21,0.49,0.74}
\usepackage[pagebackref,breaklinks,colorlinks,allcolors=cvprblue]{hyperref}

\usepackage{booktabs}
\usepackage{hyperref}
\usepackage{url}
\usepackage{xspace}
\usepackage{graphicx}
\usepackage{booktabs}
\usepackage{pifont}
\newcommand{\cmark}{\ding{51}}
\newcommand{\xmark}{\ding{55}}
\usepackage{multirow}
\usepackage{amsmath}
\usepackage[table]{xcolor}
\usepackage{colortbl}
\usepackage{tabularx}
\usepackage{array}

\def\confName{CVPR}
\def\confYear{2027}

\title{Think Before You Distill: Generalization Recovery in Quantized VLAs via Selective Policy Distillation}

\author{
Shen Ruan$^{1,*}$ \quad
Wenchang Gao$^{2,*}$ \quad
Jin Wang$^{1,\dagger}$ \quad
Siao Liu$^{1,\dagger}$ \quad
Zhaxizhuoma$^{3}$ \\
Dongchun Ren$^{4}$ \quad
Xin Zheng$^{5}$ \\[6pt]
$^{1}$Soochow University \quad
$^{2}$Independent Researcher \quad
$^{3}$Shanghai Jiao Tong University \\
$^{4}$Yootta \quad
$^{5}$Royal Melbourne Institute of Technology \\[4pt]
{\ttfamily\small ruan\_shen@163.com \quad wenchanggao93@gmail.com} \\
{\ttfamily\small ustc\_wangjin@hotmail.com \quad saliu@suda.edu.cn} \\[4pt]
{\small $^*$Equal contribution. \quad $^\dagger$Corresponding authors.}
}

\newcommand{\ours}{\textsc{PIVOT-Q}\xspace}

\begin{document}
\maketitle

\begin{abstract}
Post-training quantization has been shown to preserve VLA performance under standard evaluation conditions, but whether it preserves the full-precision model's robustness and generalization remains underexplored.
In this study, we systematically study the robustness and generalization of post-quantized VLA policies under environmental disturbances.
Empirical results show that quantized policies can become fragile to subtle environmental variations despite retaining comparable in-distribution performance. 
We further observe that action discrepancies are concentrated in a small subset of the rollout states, while teacher guidance has opposite effects depending on discrepancy: it improves performance at high-discrepancy states but can degrade it at low-discrepancy states. 
These findings reveal that effective post-quantization recovery requires selective intervention on vulnerable states rather than globally distilling the student. 
We therefore propose Policy-Induced Vulnerability-Oriented Tuning (PIVOT-Q), a vulnerability-aware On-Policy Distillation (OPD) framework that selectively corrects vulnerable states encountered during quantized-student rollouts using the frozen full-precision policy as a teacher. 
PIVOT-Q identifies vulnerable states using discounted accumulated discrepancies over a short horizon, applies phase-balanced sparse supervision, and uses a Behavioral Anchor to prevent unintended changes. 
Experiments under seven LIBERO-Plus environmental variations demonstrate consistent recovery across multiple VLA backbones and quantization methods. 
Notably, PIVOT-Q outperforms full-state distillation with smaller variance across evaluated settings while using only 20\% of its state-level distillation budget.
Our code is available at \url{https://github.com/ruanruan-andy/PIVOT-Q}.
\end{abstract}
\vspace{-8mm}

\section{Introduction}
\vspace{-2mm}
Vision-language-action (VLA) policies have demonstrated strong capabilities in language-guided robotic manipulation, yet their substantial computational and memory footprints remain a major obstacle to efficient deployment~\citep{kim2024openvla}. Post-training quantization (PTQ) provides a practical way to reduce these costs, and recent methods have shown that VLA policies can be substantially compressed while largely preserving their performance under in-distribution evaluation settings~\citep{frantar2023gptq,xiao2023smoothquant,zhang2026quantvla,xu2026qvla,wang2026holoqvla}. However, preserving performance on the original evaluation does not guarantee comparable generalization under environmental variations. To the best of our knowledge, the effect of quantization on VLA generalization has not been systematically studied.

\begin{figure}[!t]
    \centering
    \includegraphics[width=0.95\linewidth]{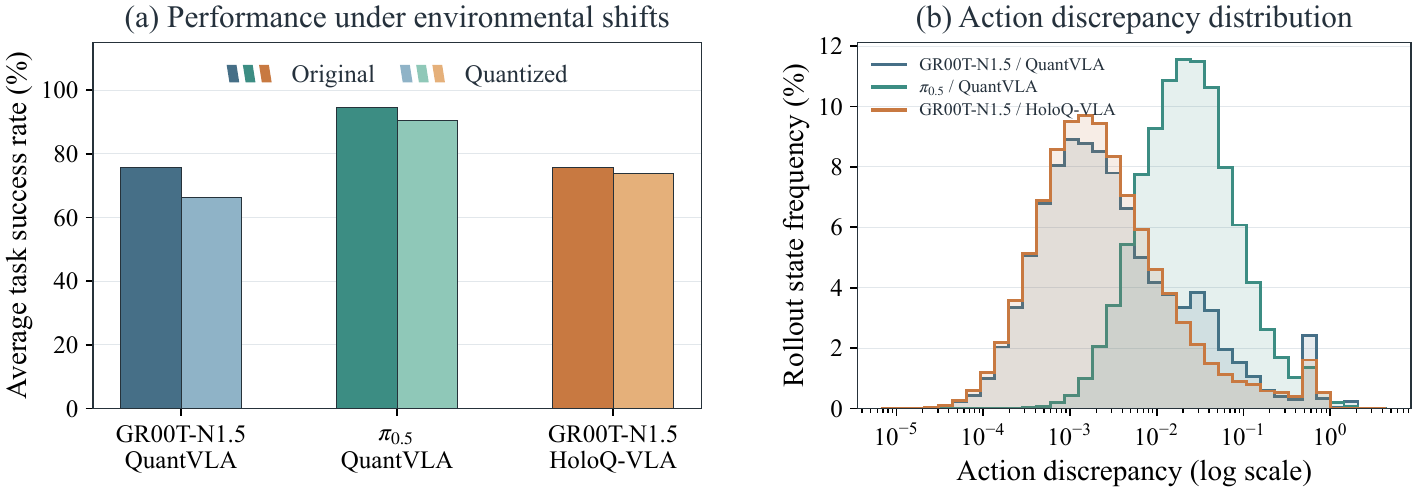}
\caption{
\textbf{Generalization degradation after quantization and action discrepancy.}
(a) Average success over LIBERO-Plus for each original and quantized policy.
(b) Distributions of action discrepancy between each quantized policy and its original counterpart.
The horizontal action discrepancy axis is in log-scale.
}
    \label{fig:gap}
\vspace{-6mm}
\end{figure}

In this work, we begin from an empirical study of VLA generalization after post-training quantization. Across multiple VLA backbones and quantization methods, we identify a consistent post-quantization generalization gap: quantized policies degrade substantially more than their full-precision counterparts under subtle disturbances in the environment, such as slight change of lighting, initial object poses, etc. 
For example, Figure~\ref{fig:gap}(a) shows that QuantVLA reduces the mean success rate of GR00T-N1.5 on LIBERO-Plus~\citep{fei2026liberoplus} from 75.7\% to 66.4\%, a decrease of 9.3 percentage points under environmental variations.
This finding raises a key question: \emph{what behavioral changes induced by quantization underlie this generalization gap?}

To investigate this question, we analyze action discrepancies between quantized and full-precision policies during rollouts under environmental disturbances. Specifically, we roll out the quantized policy and, at each visited state, compare its predicted action with that of its full-precision counterpart.
As shown in Figure~\ref{fig:gap}(b), most student-visited states exhibit small action discrepancies, while a smaller subset exhibits substantially larger disagreements.
This concentration of large discrepancies motivates prioritizing a subset of states for performance recovery.
More importantly, selectively correcting high-discrepancy states substantially restores performance, whereas correcting low-discrepancy states surprisingly leads to even worse performance. 
Effective post-quantization recovery does not require correcting every state, it simply requires correcting the right ones.

These observations suggest that uniformly distilling all student-visited states would be suboptimal. 
Although the full-precision teacher provides a stronger behavioral reference, its predictions are not guaranteed to be reliable on every state induced by the quantized student under distribution shifts. 
Moreover, in sequential robotic imitation learning, action errors differ not only in magnitude but also in their potential influence on subsequent behavior. Uniform supervision allocates adaptation capacity to all visited states, including those where correction is unnecessary or less consequential, rather than prioritizing deviations associated with persistent downstream errors. This motivates selective distillation that concentrates supervision on potentially vulnerable states while preserving already retained behaviors.

Motivated by these observations, we propose Policy-Induced Vulnerability-Oriented Tuning (\ours), a vulnerability-aware On-Policy Distillation (OPD) framework that selectively corrects sensitive states encountered by the quantized student under the guidance of the frozen full-precision teacher. Rather than uniformly distilling the student, \ours focuses supervision on states where quantization-induced deviations are most consequential, weighting their contribution according to their vulnerability. To prevent targeted recovery from disrupting behaviors already preserved after quantization, we further introduce a Behavioral Anchor that regularizes the student toward the original policy on reference states. Together, these components enable targeted correction where it matters most while preserving behaviors that are already robust to quantization.

We evaluate \ours across five model--quantizer settings spanning three VLA backbones and three quantization methods: GR00T-N1.5 and $\pi_{0.5}$ with W4A8 QuantVLA, both backbones with W4A4 HoloQ-VLA, and OpenVLA-OFT with QVLA. Extensive experiments on LIBERO-Plus cover seven controlled distribution shifts.
\ours consistently improves the corresponding quantized policies. 
It also outperforms full-state distillation in all five settings with very small variations while selecting only 20\% of student-visited states for supervision.
These results demonstrate that vulnerability-oriented sparse supervision provides consistent post-quantization recovery across different backbones, quantization schemes, and environmental shifts.

Our contributions include:
$\spadesuit$ We provide a systematic study of environmental generalization after VLA quantization, revealing a consistent generalization gap and variation in action discrepancies between the teacher and student across rollout states.
$\spadesuit$ We propose \ours, a vulnerability-aware post-quantization recovery framework that prioritizes a sparse set of student rollout states, while mitigating unintended policy drift with a Behavioral Anchor.
$\spadesuit$ We demonstrate the effectiveness of \ours across multiple VLA backbones and quantization methods. \ours consistently outperforms full-state distillation across all our evaluation settings while selecting only 20\% of student-visited states for supervision.

\vspace{-2mm}

\section{Related Work}
\vspace{-2mm}

\noindent\textbf{VLA Generalization and Quantization.}
VLA policies span discrete-action systems such as RT-2~\citep{zitkovich2023rt2} and OpenVLA~\citep{kim2024openvla}, as well as continuous generative policies such as $\pi_{0.5}$~\citep{intelligence2025pi05visionlanguageactionmodelopenworld} and GR00T~\citep{bjorck2025groot}. Their environmental robustness has been evaluated under controlled variations in viewpoint, lighting, language,  and other factors~\citep{fei2026liberoplus}. In parallel, post-training quantization reduces inference cost through low-precision weights and activations~\citep{yao2022zeroquant,yuan2023rptq,lin2024awq}. Recent VLA-specific methods, including QuantVLA~\citep{zhang2026quantvla}, QVLA~\citep{xu2026qvla}, and HoloQ-VLA~\citep{wang2026holoqvla}, further improve action preservation after quantization. However, their evaluations primarily focus on the original task distributions, leaving environmental generalization after VLA quantization largely unexplored.

\noindent\textbf{Policy Distillation and On-Policy Adaptation.} Knowledge distillation has been widely used for model compression and policy transfer~\citep{hinton2015distilling,rusu2016policy,fang2025flex}. Recent VLA methods extend this paradigm to action generation: VITA-VLA~\citep{dong2025vitavla} and ActDistill~\citep{ye2025actdistill} transfer action knowledge to lightweight policies, while VLA-OPD~\citep{zhong2026vlaopd} performs distillation on student-visited states to better match the deployment-time state distribution. Quantization-aware approaches such as QAD~\citep{xin2026qad} and SQuaT~\citep{lee2026squat} further use high-precision teachers to recover low-precision models. However, existing recovery methods do not explicitly account for how the behavioral importance of quantization errors varies across rollout states, particularly under environmental shifts.

\vspace{-2mm}

\section{Problem Formulation}
\label{sec:problem_formulation}
\vspace{-2mm}

Let $\pi^{F}$ denote a pretrained full-precision VLA policy and $\pi^{Q}$ its post-training quantized counterpart. Given an observation $o_t$, language instruction $x$, and stochastic input $\xi_t$, both policies predict an action chunk, while only the first action is executed in the environment. We denote the corresponding executed actions as $\mathbf{a}^{F}_t$ and $\mathbf{a}^{Q}_t$. When comparing the two policies, we use the same stochastic input $\xi_t$ to isolate the behavioral deviation introduced by quantization. 

In this work, we focus on post-quantization recovery under environmental variations. Since deviations in $\pi^{Q}$ can alter subsequent observations, recovery should account for the state distribution induced by the quantized policy itself rather than relying only on fixed calibration data or teacher-generated trajectories. Accordingly, we consider trajectories collected by executing $\pi^{Q}$ and measure the teacher--student action discrepancy at each visited state as:
\begin{equation}
q_t =
\frac{1}{D_a}
\left\|
\mathbf{a}^{Q}_t-\mathbf{a}^{F}_t
\right\|_2^2,
\label{eq:action_discrepancy}
\end{equation}
where $D_a$ denotes the action dimensionality. Our goal is to recover the environmental generalization of $\pi^{Q}$ using the frozen $\pi^{F}$ as a teacher, while limiting supervision to a small subset of student-visited states and avoiding unnecessary changes to behaviors already preserved after quantization. 
\vspace{-6mm}

\section{Motivation Study}
\vspace{-2mm}
Although quantization improves the deployment efficiency of VLA policies while preserving strong performance under standard evaluation conditions, whether quantized VLA policies remain robust under environmental variations has not been studied. 
We investigate this question by evaluating post-quantization environmental generalization and analyzing how quantization-induced action discrepancies are distributed across rollout states and whether they are behaviorally consequential.

\subsection{Post-Quantization Generalization Gap}
\vspace{-2mm}

We first examine the impact of quantization on model generalization by comparing the success rates of the original and quantized policies under environmental shifts. 
We consider VLA policies trained on the LIBERO suite~\citep{liu2023libero} and evaluate their post-quantization generalization on LIBERO-Plus~\citep{fei2026liberoplus}. 
By preserving the task objective while introducing controlled environmental shifts, LIBERO-Plus enables a direct comparison of the environmental robustness of full-precision and quantized policies. 

As shown in Figure~\ref{fig:gap}(a), quantization consistently reduces success rate across our evaluated models and quantization methods. 
GR00T-N1.5 with QuantVLA drops from 80.00\% to 65.36\%, $\pi_{0.5}$ with QuantVLA drops from 94.82\% to 89.82\%, and GR00T-N1.5 with HoloQ-VLA drops from 79.82\% to 74.64\%. 
These results reveal a consistent post-quantization generalization gap, suggesting that strong performance retention under standard evaluation does not imply preservation of model robustness.

\subsection{Action Discrepancy Analysis}
\vspace{-2mm}
The observed generalization gap motivates a detailed examination of how quantization changes policy behavior during execution. 
We use the quantized policies to sample new rollouts in the generalized environments again. 
At each visited state, the original and the quantized policies are provided with the same observation and diffusion noise to produce their actions. 

Figure~\ref{fig:gap}(b) summarizes the resulting action-discrepancy distributions across the evaluated model--quantizer settings. 
For each rollout, we group the log-scaled discrepancy into bins and divide the count in each bin by the total number of valid states. 
These proportions are averaged across rollouts to estimate the distribution of action discrepancy. 
As shown in Figure~\ref{fig:gap}(b), the discrepancies exhibit approximately log-normal distributions across all evaluated settings, with a pronounced right tail. 
Most visited states lie near the main concentration of the distribution, whereas substantially larger deviations occur at a smaller fraction of rollout states.

\begin{table}[!t]
\centering
\caption{\textbf{Effect of intervention at different action-discrepancy levels}
Task success rate (\%) in the GR00T-N1.5 QuantVLA setting.
``Ran. Int.'', ``HD Int.'', and ``LD Int.'' denote interventions at random,
high-discrepancy, and low-discrepancy states, respectively.
Superscripts indicate absolute changes relative to QuantVLA.
Best and second-best results among intervention strategies are shown in
\textbf{bold} and \underline{underline}, respectively.
The full-precision original policy is shaded as a reference.}
\label{tab:rescue}

\small
\setlength{\tabcolsep}{6pt}
\renewcommand{\arraystretch}{1.12}

\newcommand{\gain}[1]{\rlap{$^{\color{green!60!black}\uparrow #1}$}}
\newcommand{\loss}[1]{\rlap{$^{\color{red!70!black}\downarrow #1}$}}
\newcommand{\same}{\rlap{$^{\color{gray}0.0}$}}

\resizebox{\columnwidth}{!}{%
\begin{tabular}{l*{8}{c}}
\toprule
Method & Cam. & Init. & Lang. & Light & Back. & Noise & Layout & Mean \\
\midrule

\rowcolor{gray!12}
Original
& 85.0 & 42.5 & 78.8 & 95.0 & 96.2 & 85.0 & 88.8 & 81.6 \\

QuantVLA
& 61.2 & 30.0 & 60.0 & 92.5 & 75.0 & 61.2 & 85.0 & 66.4 \\

\midrule

Ran. Int.
& \underline{66.3}\gain{5.1}
& \textbf{33.8}\gain{3.8}
& \underline{61.3}\gain{1.3}
& \underline{93.8}\gain{1.3}
& 75.0\same
& \underline{60.0}\loss{1.2}
& \textbf{87.5}\gain{2.5}
& \underline{68.2}\gain{1.8} \\

HD Int.
& \textbf{67.5}\gain{6.3}
& 30.0\same
& \textbf{76.2}\gain{16.2}
& \textbf{96.2}\gain{3.7}
& \underline{76.2}\gain{1.2}
& \textbf{87.5}\gain{26.3}
& 82.5\loss{2.5}
& \textbf{73.8}\gain{7.4} \\

LD Int.
& 62.5\gain{1.3}
& \underline{32.5}\gain{2.5}
& 58.8\loss{1.2}
& 86.2\loss{6.3}
& \textbf{78.8}\gain{3.8}
& 58.8\loss{2.4}
& \underline{86.2}\gain{1.2}
& 66.2\loss{0.2} \\

\bottomrule
\end{tabular}}
\vspace{-4mm}
\end{table}

To evaluate whether large action discrepancies are behaviorally consequential, we conduct an intervention study on GR00T-N1.5 with QuantVLA. 
During quantized-policy rollouts, we replace the quantized action with the corresponding full-precision action at selected states, using high-discrepancy, low-discrepancy, or random states as intervention targets. 
As shown in Table~\ref{tab:rescue}, intervening at high-discrepancy states substantially improves task success. Random intervention yields a mild improvement. 
In contrast, intervening at low-discrepancy states provides no improvement and even slightly degrades performance. 
These results indicate that quantization-induced discrepancies differ in their behavioral consequences, motivating recovery to selectively prioritize consequential states rather than uniformly distill all rollout states.
\emph{Effective post-quantization recovery does not require correcting every state,  it just requires correcting the right ones}.

\vspace{-2mm}

\section{Method}
\begin{figure*}[!t]
    \centering
    \includegraphics[width=1\textwidth]{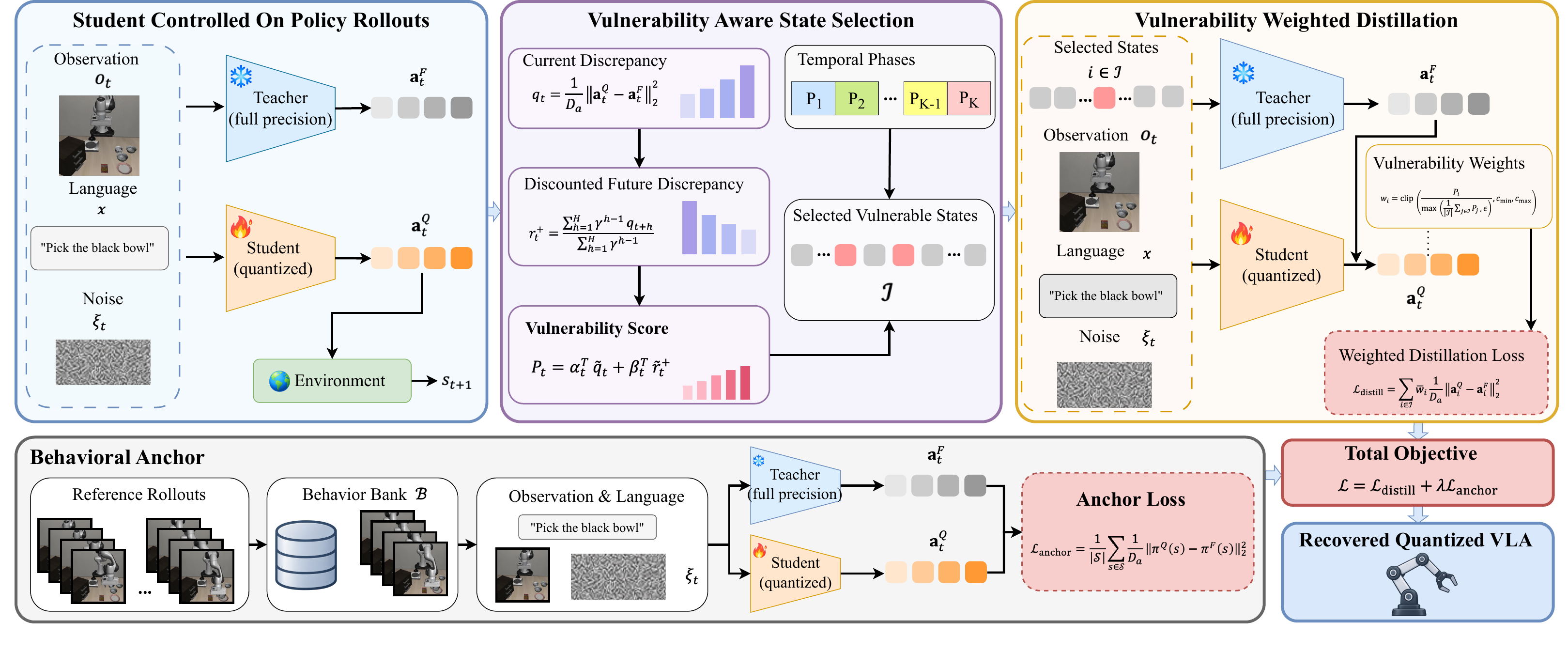}
    \vspace{-6mm}
    \caption{\textbf{Overview of \ours.} \ours identifies vulnerable states from student-controlled rollouts using accumulated discrepancy within a short future segment, performs phase-balanced sparse distillation, and uses a Behavioral Anchor to preserve retained behavior.}
    \label{fig:framework}
\vspace{-6mm}
\end{figure*}

To identify the right states for recovery, we propose Policy-Induced Vulnerability-Oriented Tuning (\ours), a vulnerability-aware OPD framework that selectively corrects vulnerable states encountered by the quantized student under the guidance of the frozen original policy. 
\ours first performs student-controlled on-policy rollouts while measuring teacher--student discrepancies along the resulting trajectories. 
It then identifies a sparse set of vulnerable states based on their discounted accumulated discrepancies over a short future horizon, with phase-wise selection of a subset of states to ensure coverage throughout the rollout.
The selected states are distilled with vulnerability-aware weighting, while a Behavioral Anchor regularizes the adapted student on reference states from the original distribution to avoid unintended drift. 
Figure~\ref{fig:framework} summarizes the overall pipeline.

\subsection{Student-Controlled On-Policy Rollouts}
\vspace{-2mm}
Fixed calibration data and teacher-generated trajectories do not explicitly track the state distribution induced by the quantized student, whose action deviations can alter subsequent observations. We therefore collect recovery trajectories on-policy using the quantized student $\pi^Q$. At each visited observation $O_t$, the student $\pi^Q$ and frozen full-precision teacher $\pi^F$ predict actions $\mathbf{a}_t^Q$ and $\mathbf{a}_t^F$ under the same diffusion noise $\xi_t$, while only $\mathbf{a}_t^Q$ is executed. This ensures that rollouts follow the student-induced state distribution, while controlling generation stochasticity so that teacher--student discrepancies provide a consistent measure of vulnerability at visited states.

\subsection{Vulnerability Aware State Selection}
\label{sec:Vulnerability Aware State Selection}
\vspace{-2mm}
The action discrepancy $q_t$ captures only the immediate teacher--student deviation. 
In sequential imitation learning, the importance of correcting a state cannot always be characterized by its instantaneous action discrepancy, as local deviations can be associated with persistent downstream behavioral errors.
Moreover, instantaneous discrepancies can fluctuate due to observation noise and stochastic action generation, making state selection sensitive to transient disagreements.

This motivates extending state prioritization beyond instantaneous discrepancy to incorporate short-horizon temporal information.
Therefore, we assess each state by accumulating future discrepancies, emphasizing nearby deviations while reducing the influence of increasingly uncertain distant steps. 
We define the discounted accumulated discrepancy over future states within a horizon $H$ as:

\begin{equation}
r_t^{+}
=
\frac{
\sum_{h=1}^{H}\gamma^{h-1}q_{t+h}
}{
\sum_{h=1}^{H}\gamma^{h-1}
},
\label{eq:future_risk}
\end{equation}
where $\gamma$ is the temporal discount factor. 
This score assigns greater importance to temporally closer steps while complementing the instantaneous discrepancy $q_t$.

To ensure temporal coverage over the whole trajectory, we partition each trajectory into $K$ contiguous phases and compute within-phase percentile ranks $\alpha^T_t, \beta^T_t$ for both discrepancy signals. 
We denote the resulting normalized ranks of the current and discounted future discrepancies as $\tilde{q}_t$ and $\tilde{r}_t^{+}$. The vulnerability score is finally defined as:
\begin{equation}
P_t
=
\alpha_t^T\,\tilde{q}_t
+
\beta_t^T\,\tilde{r}_t^{+},
\label{eq:vulnerability_score}
\end{equation}
where $\alpha_t^T$ and $\beta_t^T$ control the relative contributions of current and discounted future discrepancy within the corresponding phase and superscript $T$ denotes total steps in the phase. 
Both $\tilde{q_t}$ and $\tilde{r}_t^+$ are computed solely from teacher and student action predictions. 
No additional annotations such as rewards or task success labels are required.

Since high-vulnerability states might cluster within a short temporal interval inside each phase, directly selecting the highest-scoring states can allocate multiple supervision slots to neighboring states. We therefore greedily select states in descending order of $P_t$, enforcing a minimum temporal separation between any two selected states within the same phase.
We retain a proportion of states through phase-wise selection, resulting in a sparse supervision set $\mathcal{I}$. This constraint complements the phase-wise allocation by preventing the supervision budget from concentrating on a narrow temporal region while prioritizing states with high estimated vulnerability.

\subsection{Vulnerability Weighted Distillation}
\vspace{-2mm}
Although state selection restricts supervision to vulnerable states, the selected states can still differ substantially in their estimated vulnerability. 
Uniformly weighting them loses this information, so we assign larger distillation weights to states with higher vulnerability scores $P_i$. 
To account for variation in $P_i$ across trajectories and prevent a few states from dominating optimization, we normalize each score by the mean vulnerability over the selected set and clip the resulting ratio.
This converts the absolute vulnerability score into a bounded relative importance, emphasizing more vulnerable states while maintaining stable contributions from the remaining selected states. 
The clipped weights are renormalized over the selected set before computing the distillation objective:
\begin{equation}
\begin{aligned}
w_i &=
\operatorname{clip}\!\left(
\frac{P_i}{
\max\!\left(
\frac{1}{|\mathcal{I}|}
\sum_{j\in\mathcal{I}} P_j,
\epsilon
\right)},
c_{\min},c_{\max}
\right),\\
\bar{w}_i
&=
\frac{w_i}{\sum_{j\in\mathcal{I}}w_j}.
\end{aligned}
\label{eq:weights}
\end{equation}
Mean normalization makes the weighting comparable across trajectories with different vulnerability scales. 
Clipping is used to prevent extreme states from dominating optimization while ensuring that every selected state retains a non-negligible contribution. 

Using these weights, we distill the quantized student toward the frozen teacher at the selected states:
\begin{equation}
\mathcal{L}_{\mathrm{distill}}
=
\sum_{i\in\mathcal{I}}
\bar{w}_i
\frac{1}{D_a}
\left\|
\mathbf{a}_i^Q-\mathbf{a}_i^F
\right\|_2^2.
\label{eq:distill}
\end{equation}
This distillation objective concentrates the recovery update on states with higher estimated vulnerability while retaining sparse supervision over the trajectory.

\subsection{Behavioral Anchor}
\vspace{-2mm}
Our selective distillation focuses supervision on vulnerable states visited by the quantized student under environmental shifts. 
Updating the policy exclusively on these states may alter the policy too much towards the new environment since they provide large supervision signals, leading to instability during learning and forgetting. 
We therefore introduce a Behavioral Anchor that regularizes the adapted policy toward the behavior of the original full precision policy on reference states. 
We maintain a small behavior bank $\mathcal{B}$ that includes anchor states from the original dataset used to trained the full-precision policy. 
The student queries the frozen full-precision teacher on these anchor states to get the resulting teacher targets for optimization, as commonly conducted in continual learning to prevent forgetting~\citep{rebuffi2017icarl,lopezpaz2017gem}. 
At each optimization step, a subset $\mathcal{S}$ of anchor states is randomly sampled from the memory bank $\mathcal{B}$.
To further stabilize the training process and prevent policy drift, we integrate an auxiliary anchor loss:
\begin{equation}
\mathcal{L}_{\mathrm{anchor}}
=
\frac{1}{|\mathcal{S}|}
\sum_{s\in\mathcal{S}}
\frac{1}{D_a}
\left\|
\pi^Q(s)-\pi^F(s)
\right\|_2^2,
\label{eq:anchor}
\end{equation}

The overall recovery objective combines targeted correction with behavioral retention:
\begin{equation}
\mathcal{L}
=
\mathcal{L}_{\mathrm{distill}}
+
\lambda\mathcal{L}_{\mathrm{anchor}},
\label{eq:total_loss}
\end{equation}
where $\lambda$ is a hyper-parameter that controls the magnitude of the Behavioral Anchor. 

\vspace{-2mm}

\section{Experiments}
\vspace{-2mm}

We evaluate five model--quantizer settings: GR00T-N1.5 and $\pi_{0.5}$ with W4A8 QuantVLA, both backbones with HoloQ-VLA, and OpenVLA-OFT with QVLA.
Each setting includes an original policy, a quantized policy, a recovered policy with Full State Distillation (FSD), and recovered policy with \ours. 
FSD distills the student on all recorded rollout states. 
In contrast, \ours partitions each rollout into $K=4$ temporal phases and selects 20\% of valid rollout states using instantaneous and discounted future action discrepancies, together with vulnerability-weighted distillation and a Behavioral Anchor.
All methods follow the same LIBERO-Plus evaluation protocol, with results reported as mean $\pm$ standard deviation across three evaluation seeds. Recovery and evaluation use disjoint sets of rollout seeds while sharing benchmark task configurations and initial-state indices. This protocol measures performance recovery under environmental perturbations across separate stochastic executions. Appendix~\ref{app:evaluation_protocol} provides the detailed protocol.
\begin{table*}[!t]
\centering
\caption{\textbf{Performance recovery across models and quantization methods.}
We report task success rates (\%) under seven LIBERO-Plus distribution shifts as mean $\pm$ standard deviation, with standard deviations shown in gray.
FSD denotes \emph{Full-State Distillation}.
Boldface marks the higher mean between FSD and \ours.
$\Delta$ denotes the average success rate improvement over Quantized in percentage points.
TFLOP/s denotes estimated training throughput.}
\vspace{-2mm}
\label{tab:main}
\begingroup
\small
\setlength{\tabcolsep}{3pt}

\newcommand{\ms}[2]{\ensuremath{#1\,{\color{gray}\pm\,#2}}}
\newcommand{\bms}[2]{\ensuremath{\mathbf{#1}\,{\color{gray}\pm\,#2}}}
\newcommand{\dg}[1]{%
  \ifdim #1pt>0pt
    \textcolor{green!70!black}{\ensuremath{#1}}%
  \else
    \textcolor{red}{\ensuremath{#1}}%
  \fi
}

\resizebox{\linewidth}{!}{%
\begin{tabular}{llrrrrrrrrrr}
\toprule
Model & Method & Cam. & Init. & Lang. & Light & Back. &
Noise & Layout & Avg.$\uparrow$ & $\Delta\uparrow$ & TFLOP/s \\
\midrule

\multicolumn{12}{l}{\textit{QuantVLA (W4A8)}~\citep{zhang2026quantvla}} \\
\cmidrule(lr){1-12}

GR00T-N1.5 & Original
& \ms{73.33}{4.39} & \ms{40.83}{1.44} & \ms{72.08}{8.13}
& \ms{93.33}{8.32} & \ms{90.00}{11.92} & \ms{74.58}{4.02}
& \ms{85.83}{14.81} & \ms{75.71}{7.27}
& -- & -- \\

& Quantized
& \ms{62.50}{4.51} & \ms{35.00}{2.17} & \ms{62.92}{3.15}
& \ms{88.75}{8.66} & \ms{80.42}{4.02} & \ms{56.67}{2.60}
& \ms{78.75}{8.20} & \ms{66.43}{1.86}
& -- & -- \\

& FSD
& \bms{70.00}{5.45} & \bms{39.17}{1.44} & \ms{70.83}{1.44}
& \ms{91.25}{5.45} & \ms{94.17}{2.89} & \ms{66.25}{5.73}
& \ms{86.67}{13.37} & \ms{74.05}{2.58}
& \dg{+7.62} & 28.20 \\

\rowcolor{blue!7}
& \ours
& \ms{69.17}{5.64} & \ms{36.25}{2.17} & \bms{72.08}{3.15}
& \bms{97.50}{2.17} & \bms{95.83}{1.91} & \bms{71.67}{3.15}
& \bms{87.08}{11.55} & \bms{75.65}{1.62}
& \dg{+9.23} & 19.90 \\

\cmidrule(lr){1-12}

$\pi_{0.5}$ & Original
& \ms{97.08}{1.44} & \ms{87.50}{1.25} & \ms{88.75}{2.17}
& \ms{97.50}{1.25} & \ms{100.00}{0.00} & \ms{92.92}{1.91}
& \ms{98.75}{0.00} & \ms{94.64}{0.18}
& -- & -- \\

& Quantized
& \ms{89.17}{3.61} & \ms{84.58}{4.02} & \ms{83.33}{4.02}
& \ms{95.83}{1.91} & \ms{94.17}{4.02} & \ms{90.00}{1.25}
& \ms{95.42}{1.91} & \ms{90.36}{0.47}
& -- & -- \\

& FSD
& \bms{97.50}{0.00} & \ms{83.75}{3.31} & \ms{88.33}{0.72}
& \bms{97.50}{1.25} & \bms{99.58}{0.72} & \ms{88.33}{3.15}
& \bms{97.92}{1.91} & \bms{93.27}{0.45}
& \dg{+2.92} & 22.57 \\

\rowcolor{blue!7}
& \ours
& \ms{95.83}{3.82} & \bms{87.50}{2.50} & \bms{89.17}{3.15}
& \ms{97.08}{0.72} & \bms{99.58}{0.72} & \bms{89.58}{1.91}
& \ms{94.17}{1.91} & \bms{93.27}{0.27}
& \dg{+2.92} & 16.64 \\

\midrule
\multicolumn{12}{l}{\textit{HoloQ-VLA}~\citep{wang2026holoqvla}} \\
\cmidrule(lr){1-12}

GR00T-N1.5 & Original
& \ms{73.33}{4.39} & \ms{40.83}{1.44} & \ms{72.08}{8.13}
& \ms{93.33}{8.32} & \ms{90.00}{11.92} & \ms{74.58}{4.02}
& \ms{85.83}{14.81} & \ms{75.71}{7.27}
& -- & -- \\

& Quantized
& \ms{67.08}{5.20} & \ms{28.33}{2.60} & \ms{79.17}{2.60}
& \ms{97.50}{2.17} & \ms{92.92}{6.88} & \ms{66.25}{3.31}
& \ms{86.25}{1.25} & \ms{73.93}{1.89}
& -- & -- \\

& FSD
& \bms{78.75}{9.01} & \bms{32.92}{1.44} & \ms{79.17}{5.91}
& \ms{95.83}{5.05} & \ms{95.42}{4.02} & \bms{77.08}{2.89}
& \bms{91.25}{9.76} & \ms{78.63}{1.91}
& \dg{+4.70} & 20.63 \\

\rowcolor{blue!7}
& \ours
& \ms{75.00}{3.75} & \ms{31.67}{0.72} & \bms{82.08}{9.71}
& \bms{98.75}{0.00} & \bms{100.00}{0.00} & \ms{73.75}{3.31}
& \ms{90.83}{1.91} & \bms{78.87}{1.39}
& \dg{+4.94} & 31.15 \\

\cmidrule(lr){1-12}

$\pi_{0.5}$ & Original
& \ms{97.08}{0.72} & \ms{85.42}{3.82} & \ms{89.58}{1.91}
& \ms{98.75}{1.25} & \ms{99.58}{0.72} & \ms{91.67}{3.15}
& \ms{99.17}{0.72} & \ms{94.46}{0.18}
& -- & -- \\

& Quantized
& \ms{98.33}{1.44} & \ms{81.25}{2.17} & \ms{83.33}{1.44}
& \ms{98.33}{0.72} & \ms{98.33}{1.91} & \ms{85.42}{1.44}
& \ms{96.25}{1.25} & \ms{91.61}{0.31}
& -- & -- \\

& FSD
& \bms{98.75}{0.00} & \bms{84.58}{0.72} & \bms{85.42}{2.89}
& \bms{99.58}{0.72} & \ms{99.17}{1.44} & \ms{91.25}{1.25}
& \ms{96.25}{2.17} & \ms{93.57}{0.31}
& \dg{+1.96} & 24.89 \\

\rowcolor{blue!7}
& \ours
& \ms{97.92}{1.91} & \ms{82.50}{1.25} & \ms{85.00}{0.00}
& \bms{99.58}{0.72} & \bms{100.00}{0.00} & \bms{92.92}{1.91}
& \bms{97.92}{0.72} & \bms{93.69}{0.37}
& \dg{+2.08} & 10.75 \\

\midrule
\multicolumn{12}{l}{\textit{QVLA (Mixed-bit W-only)}~\citep{xu2026qvla}} \\
\cmidrule(lr){1-12}

OpenVLA-OFT & Original
& \ms{72.92}{5.20} & \ms{37.08}{2.60} & \ms{84.58}{1.44}
& \ms{79.58}{15.83} & \ms{85.00}{3.75} & \ms{74.17}{1.44}
& \ms{85.83}{6.17} & \ms{74.17}{3.88}
& -- & -- \\

& Quantized
& \ms{72.50}{4.51} & \ms{35.00}{2.50} & \ms{74.17}{1.91}
& \ms{45.42}{6.17} & \ms{77.50}{6.50} & \ms{70.42}{6.88}
& \ms{70.00}{11.25} & \ms{63.57}{3.26}
& -- & -- \\

& FSD
& \ms{66.25}{11.92} & \bms{36.67}{4.39} & \bms{73.75}{2.17}
& \bms{81.67}{14.49} & \ms{78.75}{3.75} & \bms{77.08}{1.91}
& \ms{68.75}{15.21} & \ms{68.99}{4.81}
& \dg{+5.42} & 35.74 \\

\rowcolor{blue!7}
& \ours
& \bms{73.75}{2.17} & \ms{32.92}{1.91} & \ms{69.17}{0.72}
& \ms{80.83}{1.91} & \bms{93.33}{2.89} & \ms{61.25}{8.66}
& \bms{73.33}{13.25} & \bms{69.23}{1.54}
& \dg{+5.65} & 39.28 \\

\bottomrule
\end{tabular}%
}
\endgroup
\vspace{-6mm}
\end{table*}

\vspace{-2mm}
\subsection{Main Results}
\vspace{-2mm}

Table~\ref{tab:main} shows that \ours provides consistent post-quantization recovery across model backbones and quantization methods. \ours improves over the corresponding policy in all five settings and achieves a higher average success rate than FSD, with low standard deviations across evaluation seeds. The gains are obtained by supervising only a small proportion of rollout states, suggesting that recovery quality depends more on identifying behaviorally consequential states than on increasing the density of state-level supervision.

\vspace{-4mm}

\paragraph{Recovery across models and quantization methods.}
\ours improves over the quantized policies across all five settings and achieves higher average success rates than FSD in most settings. For GR00T-N1.5 with QuantVLA, it raises success from $66.43\%$ to $75.65\%$, closely approaching the original policy at $75.71\%$. For GR00T-N1.5 with HoloQ-VLA, it reaches $78.87\%$, surpassing both FSD and the original policy. Moreover, \ours exhibits lower standard deviations across evaluation seeds than FSD in four settings. These results demonstrate effective recovery across different VLA backbones and quantization methods.


\begin{table}[!t]
\centering
\caption{\textbf{In-distribution performance on LIBERO.} Task success rates (\%) across four suites, reported as mean $\pm$ standard deviation over three evaluation seeds. Standard deviations are shown in gray. Boldface marks the higher mean between FSD and \ours. GR00T-N1.5 Original results are shared across the two quantization settings using eight denoising steps.}
\vspace{-2mm}
\label{tab:id}
\begingroup
\small
\setlength{\tabcolsep}{3pt}
\renewcommand{\arraystretch}{0.9}

\providecommand{\ms}[2]{\ensuremath{#1\,{\color{gray}\pm\,#2}}}
\providecommand{\bms}[2]{\ensuremath{\mathbf{#1}\,{\color{gray}\pm\,#2}}}

\resizebox{\columnwidth}{!}{%
\begin{tabular}{lrrrrr}
\toprule
Method & Spatial & Object & Goal & Long & Avg.$\uparrow$ \\
\midrule

\multicolumn{6}{l}{\textit{GR00T-N1.5 + QuantVLA (W4A8)}~\citep{zhang2026quantvla}} \\
\cmidrule(lr){1-6}
Original
& \ms{88.27}{0.46} & \ms{90.00}{1.40} & \ms{86.87}{0.50}
& \ms{73.87}{1.81} & \ms{84.75}{0.48} \\
Quantized
& \ms{93.53}{0.81} & \ms{91.13}{0.81} & \ms{83.07}{1.63}
& \ms{73.53}{0.23} & \ms{85.32}{0.56} \\
FSD
& \ms{93.07}{0.46} & \bms{93.07}{1.14} & \bms{86.47}{0.50}
& \ms{72.27}{1.42} & \bms{86.22}{0.54} \\
\rowcolor{blue!7}
\ours
& \bms{94.13}{0.58} & \ms{88.47}{1.29} & \ms{85.07}{0.42}
& \bms{73.67}{0.42} & \ms{85.33}{0.13} \\

\midrule
\multicolumn{6}{l}{\textit{$\pi_{0.5}$ + QuantVLA (W4A8)}~\citep{zhang2026quantvla}} \\
\cmidrule(lr){1-6}
Original
& \ms{98.47}{0.92} & \ms{99.53}{0.12} & \ms{96.53}{0.50}
& \ms{95.20}{0.60} & \ms{97.43}{0.46} \\
Quantized
& \ms{96.80}{0.87} & \ms{98.73}{0.31} & \ms{94.20}{1.06}
& \ms{92.53}{0.61} & \ms{95.57}{0.23} \\
FSD
& \ms{97.00}{0.20} & \bms{99.60}{0.00} & \ms{96.47}{0.23}
& \bms{92.73}{0.83} & \bms{96.45}{0.22} \\
\rowcolor{blue!7}
\ours
& \bms{97.27}{1.29} & \ms{99.07}{0.12} & \bms{96.73}{0.81}
& \ms{89.27}{1.81} & \ms{95.58}{0.24} \\

\midrule
\multicolumn{6}{l}{\textit{GR00T-N1.5 + HoloQ-VLA (W4A4)}~\citep{wang2026holoqvla}} \\
\cmidrule(lr){1-6}
Original
& \ms{88.27}{0.46} & \ms{90.00}{1.40} & \ms{86.87}{0.50}
& \ms{73.87}{1.81} & \ms{84.75}{0.48} \\
Quantized
& \ms{86.40}{0.92} & \ms{89.13}{0.12} & \ms{86.80}{1.40}
& \ms{75.27}{0.12} & \ms{84.40}{0.31} \\
FSD
& \bms{87.33}{2.14} & \bms{89.07}{0.95} & \ms{85.60}{0.80}
& \bms{74.93}{1.47} & \bms{84.23}{0.25} \\
\rowcolor{blue!7}
\ours
& \ms{86.33}{1.40} & \ms{88.53}{0.46} & \bms{86.00}{1.22}
& \ms{74.73}{1.47} & \ms{83.90}{0.85} \\

\midrule
\multicolumn{6}{l}{\textit{$\pi_{0.5}$ + HoloQ-VLA (W4A4)}~\citep{wang2026holoqvla}} \\
\cmidrule(lr){1-6}
Original
& \ms{98.47}{0.61} & \ms{99.53}{0.42} & \ms{97.00}{0.35}
& \ms{95.00}{0.69} & \ms{97.50}{0.40} \\
Quantized
& \ms{97.73}{0.50} & \ms{99.80}{0.00} & \ms{96.60}{0.80}
& \ms{97.20}{1.04} & \ms{97.83}{0.10} \\
FSD
& \bms{98.93}{0.23} & \ms{99.47}{0.31} & \ms{95.60}{0.92}
& \ms{96.53}{0.50} & \ms{97.63}{0.16} \\
\rowcolor{blue!7}
\ours
& \ms{98.13}{0.42} & \bms{99.73}{0.12} & \bms{96.00}{0.72}
& \bms{97.93}{0.50} & \bms{97.95}{0.18} \\

\midrule
\multicolumn{6}{l}{\textit{OpenVLA-OFT + QVLA (Mixed-bit W-only)}~\citep{xu2026qvla}} \\
\cmidrule(lr){1-6}
Original
& \ms{97.20}{0.35} & \ms{82.00}{0.00} & \ms{84.80}{0.20}
& \ms{58.20}{0.35} & \ms{80.55}{0.10} \\
Quantized
& \ms{94.33}{0.99} & \ms{61.40}{0.00} & \ms{73.13}{0.83}
& \ms{44.73}{0.23} & \ms{68.40}{0.13} \\
FSD
& \ms{96.20}{0.40} & \ms{61.80}{0.00} & \ms{72.07}{1.10}
& \bms{43.73}{1.29} & \ms{68.45}{0.35} \\
\rowcolor{blue!7}
\ours
& \bms{96.87}{0.58} & \bms{70.07}{0.12} & \bms{74.47}{1.70}
& \ms{38.60}{0.72} & \bms{70.00}{0.61} \\

\bottomrule
\end{tabular}%
}
\endgroup
\vspace{-8mm}
\end{table}

\begin{table*}[!t]
\vspace{-4mm}
\centering
\caption{\textbf{Component ablations and sparse selection baselines.}
We report task success rates (\%) under seven LIBERO-Plus distribution shifts
with GR00T-N1.5 and W4A8 QuantVLA.
All recovery methods use a 20\% state-selection budget.
Results are mean $\pm$ standard deviation, with standard deviations shown in gray.
$\boldsymbol{C}$ and $\boldsymbol{F}$ denote current and future discrepancy, respectively.
Boldface marks the highest average success rate among recovery methods.}
\vspace{-3mm}
\label{tab:ablation}
\begingroup
\small
\setlength{\tabcolsep}{3pt}
\renewcommand{\arraystretch}{1.08}

\newcommand{\ms}[2]{\ensuremath{#1\,{\color{gray}\pm\,#2}}}
\newcommand{\bms}[2]{\ensuremath{\mathbf{#1}\,{\color{gray}\pm\,#2}}}

\resizebox{\textwidth}{!}{%
\begin{tabular}{lllrrrrrrrr}
\toprule
Method & Selection & Anchor
& Cam. & Init. & Lang. & Light & Back. & Noise & Layout & Avg.$\uparrow$ \\
\midrule

Original & -- & --
& \ms{73.33}{4.39}
& \ms{40.83}{1.44}
& \ms{72.08}{8.13}
& \ms{93.33}{8.32}
& \ms{90.00}{11.92}
& \ms{74.58}{4.02}
& \ms{85.83}{14.81}
& \ms{75.71}{7.27} \\

Quantized & -- & --
& \ms{62.50}{4.51}
& \ms{35.00}{2.17}
& \ms{62.92}{3.15}
& \ms{88.75}{8.66}
& \ms{80.42}{4.02}
& \ms{56.67}{2.60}
& \ms{78.75}{8.20}
& \ms{66.43}{1.86} \\

\midrule

Per-Phase Random & Random & \cmark
& \ms{65.00}{6.25}
& \ms{42.50}{0.00}
& \ms{68.75}{6.25}
& \ms{99.58}{0.72}
& \ms{94.17}{1.44}
& \ms{73.33}{3.82}
& \ms{80.42}{12.33}
& \ms{74.82}{2.48} \\

Per-Phase Uniform & Uniform & \cmark
& \ms{66.67}{8.04}
& \ms{42.50}{5.73}
& \ms{71.25}{6.50}
& \ms{96.25}{3.75}
& \ms{95.83}{0.72}
& \ms{75.00}{2.17}
& \ms{77.92}{13.71}
& \ms{75.06}{3.92} \\

\midrule

w/o Future Discrepancy & $C$ only & \cmark
& \ms{71.25}{9.76}
& \ms{40.42}{2.60}
& \ms{72.08}{4.73}
& \ms{98.75}{1.25}
& \ms{93.75}{5.73}
& \ms{66.25}{3.31}
& \ms{85.00}{14.42}
& \ms{75.36}{4.65} \\

w/o Behavioral Anchor & $C+F$ & \xmark
& \ms{65.00}{6.50}
& \ms{35.00}{6.25}
& \ms{73.33}{1.91}
& \ms{96.67}{1.44}
& \ms{98.33}{1.91}
& \ms{66.67}{3.82}
& \ms{85.42}{16.60}
& \ms{74.35}{3.14} \\

\rowcolor{blue!7}
\ours & $C+F$ & \cmark
& \ms{69.17}{5.64}
& \ms{36.25}{2.17}
& \ms{72.08}{3.15}
& \ms{97.50}{2.17}
& \ms{95.83}{1.91}
& \ms{71.67}{3.15}
& \ms{87.08}{11.55}
& \bms{75.65}{1.62} \\

\bottomrule
\end{tabular}%
}
\endgroup
\vspace{-6mm}
\end{table*}

\vspace{-4mm}

\paragraph{State-selection efficiency results.}
\ours achieves recovery with a substantially smaller state-level supervision budget than FSD. It partitions each rollout into four temporal phases and selects $20\%$ of valid rollout states for supervision, whereas FSD supervises all recorded states. Despite this reduced budget, \ours exceeds FSD in four of the five evaluated settings and matches its average success rate in the remaining setting. These results demonstrate that vulnerability-oriented selection can retain or improve recovery performance with sparse supervision. The TFLOP/s measurements in Table~\ref{tab:main} describe training throughput, while the selected-state proportion measures supervision efficiency. 
\vspace{-4mm}

\paragraph{Recovery stability across tasks and methods.} Beyond improving average task success, PIVOT-Q also exhibits more consistent recovery across independent runs. As shown in Table~\ref{tab:main}, PIVOT-Q achieves a lower standard deviation than FSD in four of the five model–quantizer settings. 
Notably, the standard deviation decreases from 2.58 to 1.62 percentage points for GR00T-N1.5 with QuantVLA, and from 4.81 to 1.54 for OpenVLA-OFT with QVLA. 
The only exception is $\pi$0.5 with HoloQ-VLA, where the standard deviation increases slightly from 0.31 to 0.37. 
These results suggest that vulnerability-oriented selective distillation can improve not only average recovery performance but also the consistency of recovery across independent seeds. 

\subsection{In-Distribution Performance Preservation}
\label{sec:id_evaluation}
\vspace{-2mm}

We evaluate whether recovery under environmental variation preserves performance on the original LIBERO benchmark. As shown in Table~\ref{tab:id}, \ours maintains or improves aggregate success over the quantized policy cross the five settings, while GR00T-N1.5 with HoloQ-VLA exhibits a modest reduction of $0.50$ percentage points. 
This pattern indicates that targeted recovery does not generally come at the expense of performance under the original evaluation conditions.
Together with the LIBERO-Plus results, these findings show that selective supervision improves performance under environmental perturbations while largely retaining in-distribution capabilities.

\subsection{Ablation Studies}
\label{sec:ablation}
\vspace{-2mm}

We evaluate \ours and its controlled variants with GR00T-N1.5 under W4A8 QuantVLA on LIBERO-Plus.
All sparse recovery variants use the same 20\% state-selection budget and training protocol.
\textit{Per-Phase Random} randomly samples states within each temporal phase, whereas \textit{Per-Phase Uniform} selects evenly spaced valid states within each phase, both following the same phase-wise budget allocation as \ours.
\textit{w/o Future Discrepancy} removes the discounted future-discrepancy term from Eq.~\ref{eq:vulnerability_score} and ranks states using only the current teacher--student discrepancy.
\textit{w/o Behavioral Anchor} retains vulnerability-oriented state selection but removes the Behavioral Anchor regularizer.
Table~\ref{tab:ablation} summarizes the results.

\vspace{-4mm}

\paragraph{Phase-balanced sparsity alone is insufficient.}
Per-Phase Random and Per-Phase Uniform obtain
$74.82\pm2.48\%$ and $75.06\pm3.92\%$, respectively, whereas \ours reaches $75.65\pm1.62\%$.
Under the same 20\% state-selection budget, \ours improves the average success rate by $0.83$ and $0.60$ percentage points over random and uniform supervision, respectively.
These results suggest that temporal coverage alone does not determine recovery quality, as the choice of supervised states also matters.
Since the random and uniform variants remove vulnerability-based prioritization, this comparison evaluates vulnerability-oriented sparse supervision as a whole rather than attributing the improvement to sparsity alone.

\vspace{-4mm}

\begin{table*}[!t]
\centering
\caption{\textbf{Effect of the state-selection budget.} Task success rates (\%) under seven LIBERO-Plus distribution shifts for GR00T-N1.5 with W4A8 QuantVLA. Results are mean $\pm$ standard deviation. The quantized baseline receives no recovery supervision, while FSD supervises all recorded rollout states. Boldface marks the highest mean among the displayed recovery methods in each column, with ties both bold.}
\label{tab:selection_budget}
\vspace{-2mm}
\begingroup
\small
\setlength{\tabcolsep}{3pt}
\renewcommand{\arraystretch}{1.08}

\newcommand{\ms}[2]{\ensuremath{#1\,{\color{gray}\pm\,#2}}}
\newcommand{\bms}[2]{\ensuremath{\mathbf{#1}\,{\color{gray}\pm\,#2}}}

\resizebox{\textwidth}{!}{%
\begin{tabular}{llrrrrrrrr}
\toprule
Method & Selection ratio
& Cam. & Init. & Lang. & Light & Back.
& Noise & Layout & Avg.$\uparrow$ \\
\midrule

Quantized & 0\%
& \ms{62.50}{4.51}
& \ms{35.00}{2.17}
& \ms{62.92}{3.15}
& \ms{88.75}{8.66}
& \ms{80.42}{4.02}
& \ms{56.67}{2.60}
& \ms{78.75}{8.20}
& \ms{66.43}{1.86} \\

\midrule

\rowcolor{blue!7}
\ours & 20\%
& \ms{69.17}{5.64}
& \ms{36.25}{2.17}
& \ms{72.08}{3.15}
& \bms{97.50}{2.17}
& \bms{95.83}{1.91}
& \bms{71.67}{3.15}
& \bms{87.08}{11.55}
& \bms{75.65}{1.62} \\

\ours & 40\%
& \ms{64.58}{5.77}
& \ms{37.08}{3.15}
& \bms{75.42}{3.15}
& \bms{97.50}{1.25}
& \bms{95.83}{4.02}
& \ms{70.00}{4.51}
& \ms{83.33}{11.61}
& \ms{74.82}{2.95} \\

\ours & 60\%
& \ms{62.08}{7.32}
& \bms{39.58}{2.60}
& \ms{70.00}{4.33}
& \ms{93.33}{8.32}
& \ms{92.08}{5.05}
& \ms{70.83}{0.72}
& \ms{82.92}{14.49}
& \ms{72.98}{2.96} \\

\midrule

FSD & 100\%
& \bms{70.00}{5.45}
& \ms{39.17}{1.44}
& \ms{70.83}{1.44}
& \ms{91.25}{5.45}
& \ms{94.17}{2.89}
& \ms{66.25}{5.73}
& \ms{86.67}{13.37}
& \ms{74.05}{2.58} \\

\bottomrule
\end{tabular}%
}
\endgroup
\vspace{-6mm}
\end{table*}

\paragraph{Future discrepancy improves recovery stability.} 
Removing the discounted future-discrepancy term yields a comparable average success rate ($75.36\%$ vs. $75.65\%$), but substantially increases the standard deviation across evaluation seeds from $1.62$ to $4.65$ percentage points.
Notably, incorporating future discrepancy reduces the observed run-to-run standard deviation by approximately $65\%$, suggesting that its primary benefit lies in improving recovery consistency rather than merely increasing average performance.
While instantaneous discrepancies can be sensitive to transient action deviations, aggregating discrepancies over a short future horizon provides a temporally smoothed vulnerability estimate to limit the influence of isolated fluctuations during state selection.
These results suggest that future-aware vulnerability estimation contributes to more stable selective distillation, even when its improvement in average task success is modest.

\vspace{-4mm}

\paragraph{Behavioral Anchor preserves retained behaviors.}
Removing the Behavioral Anchor reduces the average success rate from $75.65\pm1.62\%$ to $74.35\pm3.14\%$, corresponding to a $1.30$ percentage-point decrease in mean performance and an approximately $48\%$ reduction in the observed standard deviation when the anchor is included, as measured by the reduction in standard deviation.
While vulnerability-oriented selection focuses supervision on states requiring substantial correction, the Behavioral Anchor constrains changes to behaviors already preserved after quantization.
The Behavioral Anchor addresses this risk by regularizing the student toward the full-precision policy on reference states, thereby constraining unintended policy drift during recovery training.
The improved mean performance and reduced variability are consistent with this regularization helping preserve retained capabilities during targeted recovery.
Together, these components enable \ours to achieve $75.65\%$ average success, closely matching the $75.71\%$ full-precision reference.

\vspace{-4mm}
\paragraph{Effect of the state-selection budget.}
We examine selection ratios of 20\%, 40\%, and 60\% using GR00T-N1.5 with W4A8 QuantVLA. As shown in Table~\ref{tab:selection_budget}, the 20\% variant achieves the highest average success rate among the displayed ratios at $75.65\%$, compared with $74.82\%$ and $72.98\%$ for the larger budgets. It also exceeds FSD at $74.05\%$, supporting 20\% as an effective supervision budget for this setting. These results show that a carefully selected subset of rollout states can provide sufficient supervision for effective recovery without requiring dense state-level distillation. Together with the selection-baseline ablations, they highlight the importance of both the supervision budget and the choice of states within that budget.

\subsection{State Selection Analysis}
\label{sec:selection_analysis}
\vspace{-2mm}

\begin{figure}[!t]
    \centering
    \includegraphics[width=0.8\linewidth]{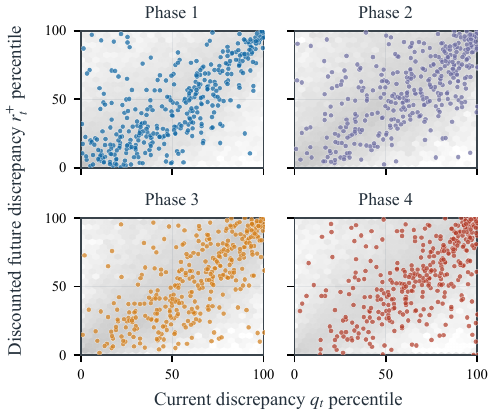}
    \vspace{-2mm}
    \caption{\textbf{Phase-wise vulnerability-oriented state selection.}
    Distribution of candidate and selected states across the four temporal
    phases for GR00T-N1.5 with W4A8 QuantVLA. Gray hexagonal bins denote
    candidate-state density, and colored points denote sampled \ours
    selections.}
    \label{fig:selection-landscape}
\vspace{-4mm}
\end{figure}

Figure~\ref{fig:selection-landscape} visualizes the states selected by \ours across the four temporal phases in the GR00T-N1.5 with W4A8 QuantVLA setting. The selected states are concentrated toward regions with high current discrepancy and high discounted future discrepancy, showing that \ours preferentially targets states with pronounced teacher--student disagreement. Meanwhile, the selected states span different combinations of the two signals rather than concentrating along a single discrepancy dimension, suggesting that current and future discrepancies provide complementary cues for identifying vulnerable states. Across all four phases, \ours consistently favors high-vulnerability regions while retaining diverse current--future discrepancy profiles. These results show that the proposed vulnerability score effectively focuses sparse supervision on behaviorally informative states throughout the rollout.

\vspace{-4mm}

\begin{figure}[!t]
    \centering
    \includegraphics[width=1\linewidth]{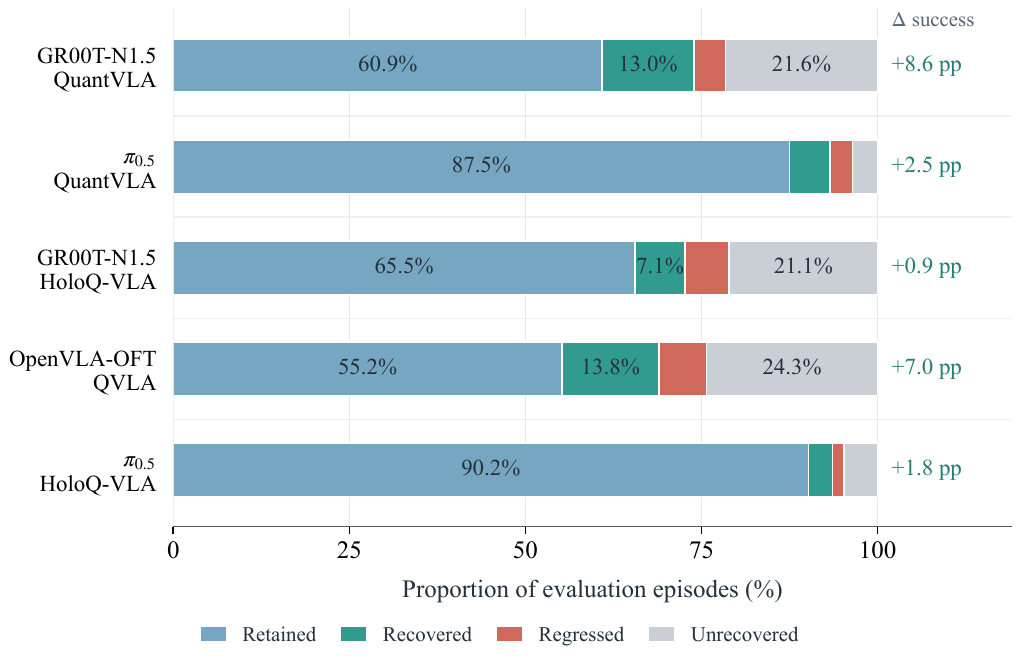}
    \vspace{-4mm}
    \caption{\textbf{Recovery and regression across five settings.} Outcome proportions for matched LIBERO-Plus episodes evaluated by the quantized policy and PIVOT-Q. Net success-rate gains are measured in percentage points and equal the recovered proportion minus the regressed proportion.}
    \label{fig:outcome_transitions}
\vspace{-4mm}
\end{figure}

\paragraph{Recovery gains outweigh regressions.}
Figure~\ref{fig:outcome_transitions} compares episode outcomes between the quantized policy and \ours using a single evaluation seed. Retained episodes succeed under both policies, recovered episodes fail under the quantized policy but succeed under \ours, regressed episodes succeed under the quantized policy but fail under \ours, and unrecovered episodes fail under both policies. Recovered episodes outnumber regressed episodes across all five settings, yielding net improvements ranging from 0.9 to 8.6 percentage points. These results show that selective supervision preserves many existing successes while recovering additional failures, although regressions and unrecovered episodes remain. The aggregate gains therefore reflect successful recovery outweighing behavioral regressions, rather than uniformly improved performance on every episode.

\vspace{-2mm}

\section{Conclusion}
\vspace{-2mm}

We systematically study environmental generalization after VLA quantization and characterize performance degradation across different VLA backbones and quantization methods. Our analysis reveals heterogeneous teacher--student action discrepancies across rollout states, motivating selective allocation of teacher supervision. We propose \ours, a vulnerability-aware on-policy distillation framework combining current and future discrepancy prioritization, phase-balanced sparse selection, and behavioral anchoring. Across five model--quantizer settings, \ours improves the quantized policies and exceeds full-state distillation in average success rate in most settings while supervising only $20\%$ of valid rollout states. These results highlight that effective post-quantization recovery depends not on supervising more states, but on identifying and prioritizing states that benefit from teacher guidance.

\newpage

\section*{LLM Usage Statement}

We used OpenAI ChatGPT and Codex to assist with literature discovery, discussions of problem formulation and experimental design, code development and debugging, experiment orchestration, figure-formatting and LaTeX code, and manuscript drafting and revision. These tools helped identify candidate references, develop and troubleshoot training and evaluation utilities, and improve the clarity and organization of the manuscript. The authors reviewed the AI-assisted text and code, checked cited references against their original sources, and verified reported numerical results against experimental records. All scientific decisions and interpretations remain those of the authors, who take full responsibility for the correctness, originality, and reproducibility of the work.

\subsection*{Reproducibility Statement}

We describe the method and experimental protocols throughout the main paper and appendices. Section~5 presents the complete PIVOT-Q framework and training objective, including student-controlled on-policy rollouts, vulnerability-aware state selection, vulnerability-weighted distillation, and the Behavioral Anchor. Section~6 specifies the evaluated model--quantizer settings, baselines, evaluation protocols, and reporting procedures, and presents performance recovery under environmental perturbations, in-distribution performance, component ablations, and the state-selection budget study. Appendix~\ref{app:implementation} provides implementation details and hyperparameter settings. Appendix~\ref{app:evaluation_protocol} specifies task selection, shared benchmark configurations, and the use of disjoint rollout-seed sets for recovery and evaluation. Appendix~\ref{app:score_sensitivity} examines sensitivity to the vulnerability-score mixing weights. Appendix~\ref{app:phase_balanced_selection} visualizes vulnerability scores and phase-balanced state selection across seven perturbation categories. We will release our training and evaluation code, experiment configurations, task manifests, and analysis scripts to support reproducibility.

\clearpage

{
    \small
    \bibliographystyle{ieeenat_fullname}
    \bibliography{main}

@inproceedings{zitkovich2023rt2,
  title     = {{RT-2}: Vision-Language-Action Models Transfer Web Knowledge to Robotic Control},
  author    = {Zitkovich, Brianna and others},
  booktitle = {Proceedings of the 7th Conference on Robot Learning},
  series    = {Proceedings of Machine Learning Research},
  volume    = {229},
  year      = {2023}
}

@inproceedings{kim2024openvla,
  title     = {{OpenVLA}: An Open-Source Vision-Language-Action Model},
  author    = {Kim, Moo Jin and Pertsch, Karl and Karamcheti, Siddharth and Xiao, Ted and Balakrishna, Ashwin and Nair, Suraj and Rafailov, Rafael and Foster, Ethan P. and Sanketi, Pannag R. and Vuong, Quan and Kollar, Thomas and Burchfiel, Benjamin and Tedrake, Russ and Sadigh, Dorsa and Levine, Sergey and Liang, Percy and Finn, Chelsea},
  booktitle = {Proceedings of the 8th Conference on Robot Learning},
  volume    = {270},
  pages     = {2679--2713},
  year      = {2025}
}

@misc{intelligence2025pi05visionlanguageactionmodelopenworld,
      title={$\pi_{0.5}$: a Vision-Language-Action Model with Open-World Generalization}, 
      author={Physical Intelligence and Kevin Black and Noah Brown and James Darpinian and Karan Dhabalia and Danny Driess and Adnan Esmail and Michael Equi and Chelsea Finn and Niccolo Fusai and Manuel Y. Galliker and Dibya Ghosh and Lachy Groom and Karol Hausman and Brian Ichter and Szymon Jakubczak and Tim Jones and Liyiming Ke and Devin LeBlanc and Sergey Levine and Adrian Li-Bell and Mohith Mothukuri and Suraj Nair and Karl Pertsch and Allen Z. Ren and Lucy Xiaoyang Shi and Laura Smith and Jost Tobias Springenberg and Kyle Stachowicz and James Tanner and Quan Vuong and Homer Walke and Anna Walling and Haohuan Wang and Lili Yu and Ury Zhilinsky},
      year={2025},
      eprint={2504.16054},
      archivePrefix={arXiv},
      primaryClass={cs.LG},
      url={https://arxiv.org/abs/2504.16054}, 
}

@article{bjorck2025groot,
  title   = {{GR00T N1}: An Open Foundation Model for Generalist Humanoid Robots},
  author  = {Bjorck, Johan and others},
  journal = {arXiv preprint arXiv:2503.14734},
  year    = {2025}
}

@inproceedings{liu2023libero,
  title     = {{LIBERO}: Benchmarking Knowledge Transfer for Lifelong Robot Learning},
  author    = {Liu, Bo and Zhu, Yifeng and Gao, Chongkai and Feng, Yihao and Liu, Qiang and Zhu, Yuke and Stone, Peter},
  booktitle = {Advances in Neural Information Processing Systems Datasets and Benchmarks Track},
  volume    = {36},
  year      = {2023}
}

@inproceedings{fei2026liberoplus,
  title     = {{LIBERO-Plus}: A Progressive Robustness Benchmark for Visual-Language-Action Models},
  author    = {Fei, Senyu and Wang, Siyin and Shi, Junhao and Dai, Zihao and Cai, Jikun and Qian, Pengfang and Ji, Li and He, Xinzhe and Zhang, Shiduo and Fei, Zhaoye and Fu, Jinlan and Gong, Jingjing and Qiu, Xipeng},
  booktitle = {Proceedings of the IEEE/CVF Conference on Computer Vision and Pattern Recognition},
  pages     = {38574--38583},
  year      = {2026}
}

@inproceedings{zhang2026quantvla,
  title     = {{QuantVLA}: Scale-Calibrated Post-Training Quantization for Vision-Language-Action Models},
  author    = {Zhang, Jingxuan and Hsieh, Yunta and Wan, Zhongwei and Lin, Haokun and Wang, Xin and Wang, Ziqi and Lei, Yingtie and Zhang, Mi},
  booktitle = {Proceedings of the IEEE/CVF Conference on Computer Vision and Pattern Recognition},
  pages     = {39539--39549},
  year      = {2026}
}

@inproceedings{xu2026qvla,
  title     = {{QVLA}: Not All Channels Are Equal in Vision-Language-Action Model's Quantization},
  author    = {Xu, Yuhao and Yang, Yantai and Fan, Zhenyang and Liu, Yufan and Li, Yuming and Li, Bing and Zhang, Zhipeng},
  booktitle = {International Conference on Learning Representations},
  year      = {2026}
}

@article{wang2026holoqvla,
  title   = {{HoloQ-VLA}: Uniform {W4A4} Quantization of Vision-Language-Action Models},
  author  = {Wang, Xinyu and Li, Mingze and Lyu, Sicheng and Liu, Dongxiu and Yang, Kaicheng and Zhao, Ziyu and Cui, Yufei and Chang, Xiao-Wen and Lu, Peng},
  journal = {arXiv preprint arXiv:2605.28803},
  year    = {2026},
  url     = {https://arxiv.org/abs/2605.28803v3}
}

@inproceedings{frantar2023gptq,
  title     = {{GPTQ}: Accurate Post-Training Quantization for Generative Pre-trained Transformers},
  author    = {Frantar, Elias and Ashkboos, Saleh and Hoefler, Torsten and Alistarh, Dan},
  booktitle = {International Conference on Learning Representations},
  year      = {2023}
}

@inproceedings{xiao2023smoothquant,
  title     = {{SmoothQuant}: Accurate and Efficient Post-Training Quantization for Large Language Models},
  author    = {Xiao, Guangxuan and Lin, Ji and Seznec, Mickael and Wu, Hao and Demouth, Julien and Han, Song},
  booktitle = {Proceedings of the 40th International Conference on Machine Learning},
  volume    = {202},
  pages     = {38087--38099},
  year      = {2023}
}

@article{hinton2015distilling,
  title   = {Distilling the Knowledge in a Neural Network},
  author  = {Hinton, Geoffrey and Vinyals, Oriol and Dean, Jeff},
  journal = {arXiv preprint arXiv:1503.02531},
  year    = {2015}
}

@inproceedings{rusu2016policy,
  title     = {Policy Distillation},
  author    = {Rusu, Andrei A. and Gomez Colmenarejo, Sergio and Gulcehre, Caglar and Desjardins, Guillaume and Kirkpatrick, James and Pascanu, Razvan and Mnih, Volodymyr and Kavukcuoglu, Koray and Hadsell, Raia},
  booktitle = {International Conference on Learning Representations},
  year      = {2016}
}

@article{dong2025vitavla,
  title={{VITA-VLA}: Efficiently Teaching Vision-Language Models to Act via Action Expert Distillation},
  author={Dong, Shaoqi and Fu, Chaoyou and Gao, Haihan and Zhang, Yi-Fan and Yan, Chi and Wu, Chu and Liu, Xiaoyu and Shen, Yunhang and Huo, Jing and Jiang, Deqiang and Cao, Haoyu and Gao, Yang and Sun, Xing and He, Ran and Shan, Caifeng},
  journal={arXiv preprint arXiv:2510.09607},
  year={2025}
}

@article{ye2025actdistill,
  title={{ActDistill}: General Action-Guided Self-Derived Distillation for Efficient Vision-Language-Action Models},
  author={Ye, Wencheng and Wang, Tianshi and Zhu, Lei and Li, Fengling and Yang, Guoli and Shen, Hengtao},
  journal={arXiv preprint arXiv:2511.18082},
  year={2025}
}

@article{zhong2026vlaopd,
  title={{VLA-OPD}: Bridging Offline SFT and Online RL for Vision-Language-Action Models via On-Policy Distillation},
  author={Zhong, Zhide and Yan, Haodong and Li, Junfeng and He, Junjie and Zhang, Tianran and Li, Haoang},
  journal={arXiv preprint arXiv:2603.26666},
  year={2026}
}

@article{xin2026qad,
  title={Quantization-Aware Distillation for {NVFP4} Inference Accuracy Recovery},
  author={Xin, Meng and Priyadarshi, Sweta and Xin, Jingyu and Kartal, Bilal and
          Vavre, Aditya and Thekkumpate, Asma Kuriparambil and Chen, Zijia and
          Mahabaleshwarkar, Ameya Sunil and Shahaf, Ido and Bercovich, Akhiad and
          Patel, Kinjal and Velury, Suguna Varshini and Luo, Chenjie and Cheng, Zhiyu and
          Chen, Jenny and Yu, Chen-Han and Ping, Wei and Rybakov, Oleg and
          Tajbakhsh, Nima and Olabiyi, Oluwatobi and Stosic, Dusan and Wu, Di and
          Han, Song and Chung, Eric and Sreenivas, Sharath Turuvekere and
          Catanzaro, Bryan and Suhara, Yoshi and Blankevoort, Tijmen and Mao, Huizi},
  journal={arXiv preprint arXiv:2601.20088},
  year={2026}
}

@inproceedings{lee2026squat,
  title={{SQuaT}: Self-Supervised Knowledge Distillation via Student-Aware Quantized Teacher Features},
  author={Lee, HyeonJun and Jo, Hyeonsik and Chung, Jinwoo and Kim, Jangho},
  booktitle={Proceedings of the 29th International Conference on Artificial Intelligence and Statistics},
  series={Proceedings of Machine Learning Research},
  volume={300},
  pages={5023--5031},
  year={2026},
  publisher={PMLR}
}

@inproceedings{yao2022zeroquant,
  title     = {{ZeroQuant}: Efficient and Affordable Post-Training Quantization for Large-Scale Transformers},
  author    = {Yao, Zhewei and Aminabadi, Reza Yazdani and Zhang, Minjia and Wu, Xiaoxia and Li, Conglong and He, Yuxiong},
  booktitle = {Advances in Neural Information Processing Systems},
  volume    = {35},
  year      = {2022}
}

@article{yuan2023rptq,
  title   = {{RPTQ}: Reorder-based Post-training Quantization for Large Language Models},
  author  = {Yuan, Zhihang and Niu, Lin and Liu, Jiawei and Liu, Wenyu and Wang, Xinggang and Shang, Yuzhang and Sun, Guangyu and Wu, Qiang and Wu, Jiaxiang and Wu, Bingzhe},
  journal = {arXiv preprint arXiv:2304.01089},
  year    = {2023}
}

@inproceedings{lin2024awq,
  title     = {{AWQ}: Activation-aware Weight Quantization for On-Device LLM Compression and Acceleration},
  author    = {Lin, Ji and Tang, Jiaming and Tang, Haotian and Yang, Shang and Chen, Wei-Ming and Wang, Wei-Chen and Xiao, Guangxuan and Dang, Xingyu and Gan, Chuang and Han, Song},
  booktitle = {Proceedings of Machine Learning and Systems},
  volume    = {6},
  year      = {2024}
}

@inproceedings{rebuffi2017icarl,
  title     = {iCaRL: Incremental Classifier and Representation Learning},
  author    = {Rebuffi, Sylvestre-Alvise and Kolesnikov, Alexander and Sperl, Georg and Lampert, Christoph H.},
  booktitle = {Proceedings of the IEEE Conference on Computer Vision and Pattern Recognition},
  pages     = {2001--2010},
  year      = {2017}
}

@inproceedings{lopezpaz2017gem,
  title     = {Gradient Episodic Memory for Continual Learning},
  author    = {Lopez-Paz, David and Ranzato, Marc'Aurelio},
  booktitle = {Advances in Neural Information Processing Systems},
  volume    = {30},
  year      = {2017}
}

@misc{fang2025flex,
      title={FLEX: A Framework for Learning Robot-Agnostic Force-based Skills Involving Sustained Contact Object Manipulation}, 
      author={Shijie Fang and Wenchang Gao and Shivam Goel and Christopher Thierauf and Matthias Scheutz and Jivko Sinapov},
      year={2025},
      eprint={2503.13418},
      archivePrefix={arXiv},
      primaryClass={cs.RO},
      url={https://arxiv.org/abs/2503.13418}, 
}
}
\clearpage
\newpage

\appendix

\section{Implementation Details}
\label{app:implementation}

Table~\ref{tab:hyperparameters} summarizes the recovery hyperparameters and configurations across the five evaluated settings. We adapt the quantized student using LoRA while keeping the full-precision teacher frozen. Each round collects one student-controlled rollout, followed by optimization on the selected states. The number of optimizer updates depends on the number of selected states and the training batch size rather than a fixed update count.

For vulnerability estimation, we combine the current discrepancy with discounted future discrepancies using percentile ranks computed within each temporal phase. Each trajectory is partitioned into four phases, with a total selection budget corresponding to $20\%$ of valid rollout states. States are selected in descending order of vulnerability, subject to a minimum temporal separation within each phase. The selected states receive normalized vulnerability weights with clipping.

The Behavioral Anchor maintains a buffer of up to 256 clean reference states per rollout stream. At each optimizer update, up to four states are randomly sampled from each buffer, using all available states when fewer than four are stored. Samples from multiple rollout streams are combined to compute the anchor loss. FSD supervises all recorded rollout states without the Behavioral Anchor, while the no-anchor ablation retains sparse selection and disables only the anchor loss.

GR00T-N1.5 and OpenVLA-OFT use separate recovery runs for each suite, with 140 rollout episodes per suite. Each $\pi_{0.5}$ setting instead uses a shared recovery run across the four suites, totaling 560 rollout episodes. The separation between recovery and evaluation is specified in Appendix~\ref{app:evaluation_protocol}.

\section{Evaluation Protocol}
\label{app:evaluation_protocol}

For each LIBERO-Plus suite, Spatial, Object, Goal, and Long, we select 20 benchmark task entries from each of the seven perturbation categories according to the predefined ordering recorded in our task manifest. This yields 140 evaluation episodes per suite and 560 episodes per evaluation seed. The selected task entries, initial-state indices, and evaluation seeds are held fixed across methods.

\paragraph{Separation between recovery and evaluation.}
Recovery uses rollout seed $0$, whereas evaluation uses a disjoint set of rollout seeds. Evaluation episodes are collected through separate policy executions rather than by scoring stored recovery trajectories. The separation is therefore defined at the rollout-seed level. Benchmark task configurations, perturbation categories, and saved initial-state indices are shared between recovery and evaluation. Changing the rollout seed controls stochastic policy and environment processes but does not establish disjoint task or perturbation configurations. Accordingly, the reported results characterize recovery on the adapted benchmark configurations across stochastic executions, rather than transfer to held-out tasks or perturbation configurations.

\section{Sensitivity to Vulnerability Score Weighting}
\label{app:score_sensitivity}

\begin{table}[!htbp]
\centering
\caption{\textbf{Sensitivity to vulnerability-score weighting.} Single-seed task success rate (\%) on 560 LIBERO-Plus episodes. Bold and underline denote the highest and second-highest distinct values in each column, including ties.}
\label{tab:score_sensitivity}
\begingroup
\footnotesize
\setlength{\tabcolsep}{3pt}
\renewcommand{\arraystretch}{1.08}
\begin{tabular}{@{}lrrrrrrrr@{}}
\toprule
\multicolumn{9}{@{}l}{\textit{GR00T-N1.5 / QuantVLA (W4A8)}} \\
\midrule
$(\alpha,\beta)$ & Cam. & Init. & Lang. & Light & Back. & Noise & Layout & Avg.$\uparrow$ \\
\midrule
$(0.3,0.7)$
& 73.8
& \textbf{40.0}
& \textbf{81.2}
& \underline{98.8}
& \textbf{100.0}
& 82.5
& \underline{88.8}
& \underline{80.7} \\

$(0.4,0.6)$
& \underline{78.8}
& 27.5
& 76.2
& \textbf{100.0}
& \textbf{100.0}
& \underline{86.2}
& 85.0
& 79.1 \\

\rowcolor{blue!7}
$(0.5,0.5)$
& \textbf{80.0}
& \underline{37.5}
& 77.5
& \underline{98.8}
& \textbf{100.0}
& \textbf{88.8}
& \textbf{90.0}
& \textbf{81.8} \\

$(0.7,0.3)$
& 76.2
& 30.0
& \underline{80.0}
& \textbf{100.0}
& \underline{97.5}
& 82.5
& 85.0
& 78.8 \\
\bottomrule
\end{tabular}
\endgroup
\end{table}

We study the sensitivity of PIVOT-Q to the relative contributions of current and discounted future discrepancy in the vulnerability score. For a state at time $t$, the score is defined as
\begin{equation}
V_t=\alpha C_t+\beta F_t,
\qquad \alpha+\beta=1,
\label{eq:appendix_score_weighting}
\end{equation}
where $C_t$ and $F_t$ denote the normalized current and discounted future discrepancy components used by the selection rule, respectively.

All experiments use GR00T-N1.5 with W4A8 QuantVLA, the same 560-episode LIBERO-Plus evaluation protocol and the same training configuration. The default $(\alpha,\beta)=(0.5,0.5)$ configuration corresponds to the canonical PIVOT-Q run. The remaining coefficient settings are single-run sensitivity experiments; they are intended to characterize the effect of score weighting rather than to establish statistically significant differences between coefficients.

The default balanced mixture obtains the highest average success rate among the tested configurations, at 81.8\%. The future-weighted $(0.3,0.7)$ mixture achieves 80.7\%, while $(0.4,0.6)$ and the current-weighted $(0.7,0.3)$ mixture achieve 79.1\% and 78.8\%, respectively. Performance therefore does not vary monotonically with either coefficient in this sweep. These results support the balanced mixture as a reasonable default for the evaluated setting. Since the coefficient sweep is based on single-run evaluations, we treat it as a sensitivity analysis rather than evidence for a uniquely optimal weighting.

\section{Vulnerability Scores and Temporal Coverage}
\label{app:phase_balanced_selection}

\begin{figure}[!t]
    \centering
    \includegraphics[width=\linewidth]{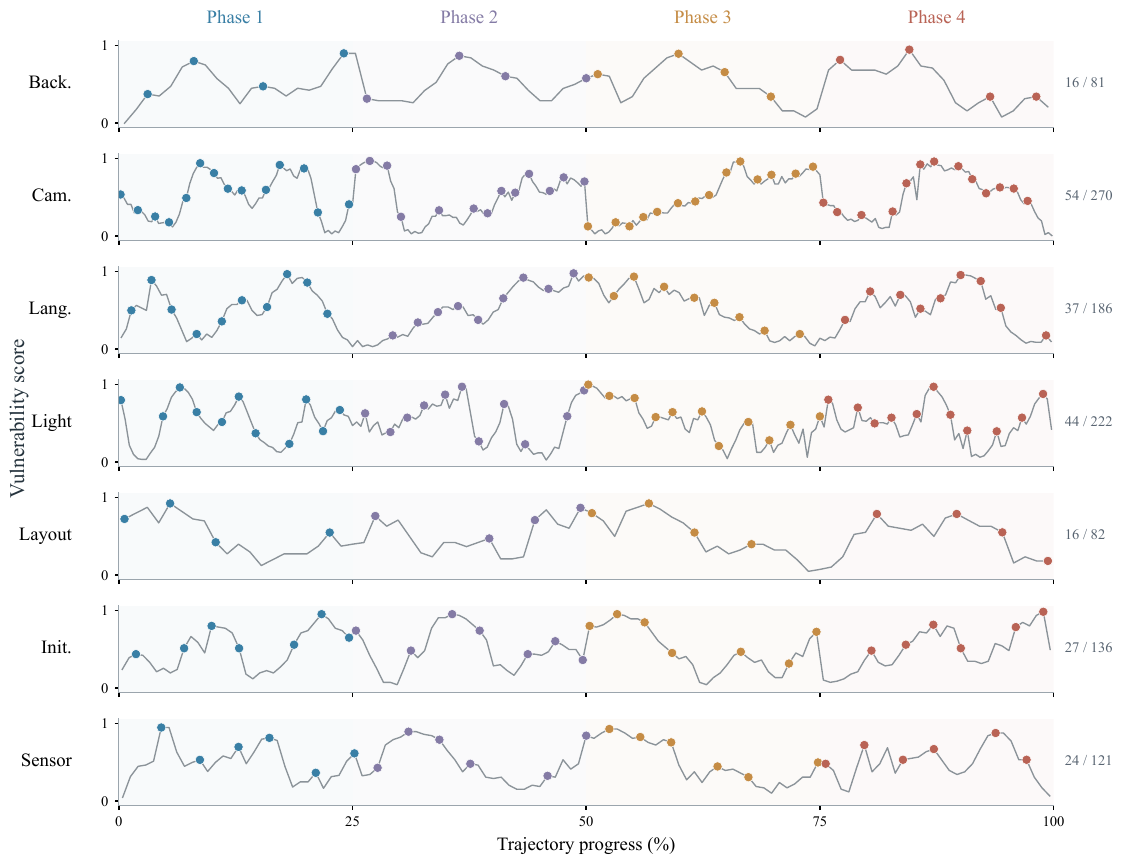}
    \caption{\textbf{Vulnerability scores and phase-balanced state selection.} Each row shows one example rollout from a perturbation category. Gray curves represent vulnerability scores, and colored points indicate selected states. Background shading and point colors distinguish the four temporal phases. The horizontal axis shows normalized trajectory progress. Counts on the right report selected states and total valid states.}
    \label{fig:phase_balanced_selection}
\end{figure}

Figure~\ref{fig:phase_balanced_selection} visualizes vulnerability scores and selected states for one example rollout from each of the seven perturbation categories. The score profiles vary across trajectories, with high-vulnerability regions occurring at different stages of execution. Selected states span all four temporal phases while prioritizing higher-scoring candidates within each phase. Selection does not simply retain global score maxima, as phase-wise budget allocation and temporal-spacing constraints also determine which states receive supervision. Consequently, some selected states have lower scores than unselected states elsewhere in the trajectory. The number of selected states also scales with trajectory length, approximately following the 20\% budget. These examples illustrate how vulnerability prioritization and temporal coverage jointly determine the sparse supervision set.

\section{Action Discrepancy and Recovery Outcomes}
\label{app:outcome_discrepancy}

\begin{figure}[!h]
    \centering
    \includegraphics[width=\linewidth]{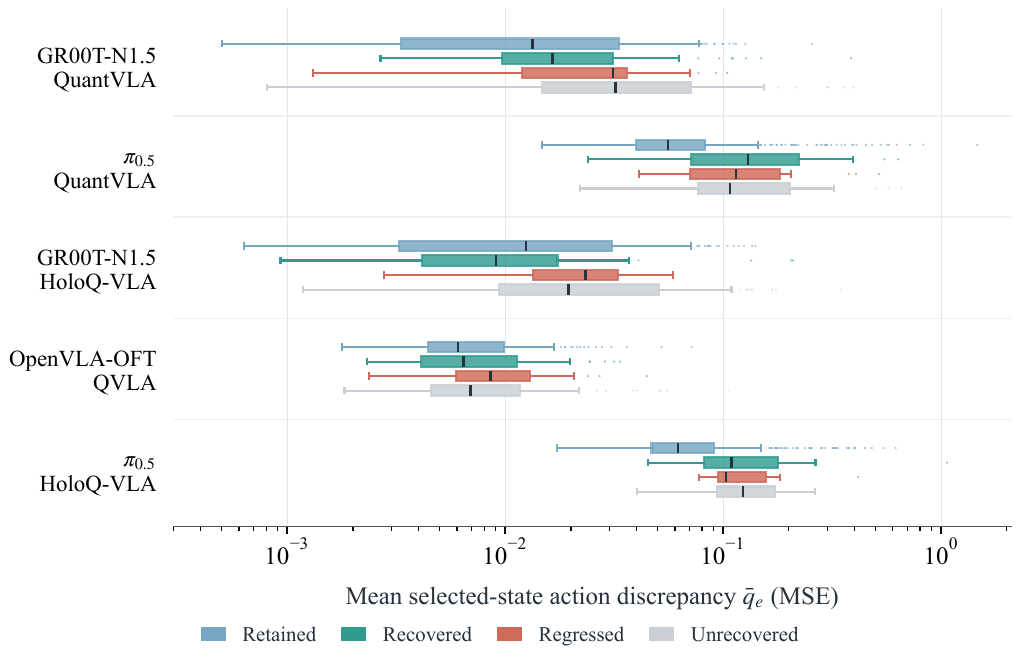}
    \caption{\textbf{Action discrepancy grouped by evaluation outcome.} Mean action discrepancies of states selected during recovery training are grouped by the corresponding final evaluation outcomes. Boxes indicate the median and interquartile range, whiskers extend to 1.5 times the interquartile range, and points beyond the whiskers denote outliers. The horizontal axis uses a logarithmic scale.}
    \label{fig:outcome_discrepancy}
\end{figure}

We examine the relationship between action discrepancy during recovery training and final evaluation outcomes across the five model and quantizer settings. For each matched evaluation episode, we compute the mean action discrepancy of the states selected during recovery training and group the resulting values by the outcome categories defined in the main text.

Figure~\ref{fig:outcome_discrepancy} reveals a tendency for recovered episodes to exhibit lower action discrepancies at the states selected during adaptation, whereas regressed episodes often exhibit larger discrepancies. However, the distributions overlap, indicating that discrepancy magnitude alone does not clearly distinguish recovery from regression. This observation motivates considering additional temporal information when prioritizing states for supervision.

\begin{table*}[!t]
\centering
\caption{\textbf{Hyperparameters and experimental configurations.} Shared recovery hyperparameters and configurations across the five model--quantizer settings.}
\label{tab:hyperparameters}
\begingroup
\scriptsize
\setlength{\tabcolsep}{4pt}
\renewcommand{\arraystretch}{1.12}

\begin{tabularx}{\textwidth}{@{}lX@{}}
\toprule
Hyperparameter & Configuration \\
\midrule
\multicolumn{2}{@{}l}{\textit{Optimization and adaptation}} \\
\midrule
Rollouts per round & $1$ \\
Learning rate & $5\times10^{-5}$ \\
Weight decay & $0.01$ \\
Gradient clipping norm & $1.0$ \\
LoRA rank & $16$ \\
LoRA scaling factor & $32$ \\
LoRA dropout & $0.05$ \\
Teacher parameters & Frozen \\
Trainable parameters & LoRA adapters \\

\midrule
\multicolumn{2}{@{}l}{\textit{Vulnerability estimation and state selection}} \\
\midrule
Future horizon $H$ & $4$ \\
Discount factor $\gamma$ & $0.9$ \\
Current discrepancy weight $\alpha$ & $0.5$ \\
Future discrepancy weight $\beta$ & $0.5$ \\
Discrepancy normalization & Within-phase percentile ranks \\
Number of temporal phases $K$ & $4$ \\
State-selection ratio & $20\%$ \\
Minimum temporal separation & $4$ rollout timesteps within each phase \\
Selection priority & Descending vulnerability score within each phase \\
Distillation weighting & Normalized vulnerability weights \\
Weight clipping range $[c_{\min},c_{\max}]$ & $[0.5,2.0]$ \\

\midrule
\multicolumn{2}{@{}l}{\textit{Behavioral Anchor}} \\
\midrule
Anchor coefficient $\lambda$ & $0.1$ \\
Reference-state buffer capacity & $256$ states \\
Reference states per optimizer update & Up to $4$ \\
Clean reference rollout horizon & $4$ \\
Reference-state source & Clean reference environments \\

\midrule
\multicolumn{2}{@{}l}{\textit{Baselines and controlled variants}} \\
\midrule
FSD & All recorded rollout states, with $\lambda=0$ \\
Per-Phase Random & Random selection within each phase using the same $20\%$ budget \\
Per-Phase Uniform & Uniform temporal selection within each phase using the same $20\%$ budget \\
w/o Future Discrepancy & Current discrepancy only, retaining the remaining components \\
w/o Behavioral Anchor & Original selection and weighting, with $\lambda=0$ \\
Additional selection ratios & $40\%$ and $60\%$, retaining the remaining recovery parameters \\
\bottomrule
\end{tabularx}

\vspace{2mm}

\begin{tabularx}{\textwidth}{@{}l*{5}{>{\raggedright\arraybackslash}X}@{}}
\toprule
Configuration
& GR00T-N1.5 + QuantVLA
& $\pi_{0.5}$ + QuantVLA
& GR00T-N1.5 + HoloQ-VLA
& $\pi_{0.5}$ + HoloQ-VLA
& OpenVLA-OFT + QVLA \\
\midrule

Quantization
& W4A8
& W4A8
& W4A4
& W4A4
& Mixed-bit weight-only \\

PIVOT-Q batch size
& $32$
& $32$
& $32$
& $16$
& $32$ \\

FSD batch size
& $32$
& $32$
& $32$
& $16$
& $32$ \\

Recovery rollout budget
& $140$ per suite
& $560$ across four suites
& $140$ per suite
& $560$ across four suites
& $140$ per suite \\

Adapter organization
& Separate per suite
& Shared across suites
& Separate per suite
& Shared across suites
& Separate per suite \\

Denoising/flow steps
& $8$
& $10$
& $8$
& $10$
& -- \\

Action execution steps
& $1$
& $1$
& $1$
& $1$
& $1$ \\

\bottomrule
\end{tabularx}

\endgroup
\end{table*}

\end{document}